\documentclass{article}

\usepackage[preprint]{neurips_2025}

\usepackage[utf8]{inputenc} 
\usepackage[T1]{fontenc}    
\usepackage{hyperref}       
\usepackage{url}            
\usepackage{booktabs}       
\usepackage{amsfonts}       
\usepackage{nicefrac}       
\usepackage{microtype}      
\usepackage{xcolor}         

\usepackage{graphicx}
\usepackage{amsmath}
\usepackage{multirow} 
\usepackage{algorithm}
\usepackage{algpseudocode}

\title{FlowCut: Unsupervised Video Instance\\ Segmentation via Temporal Mask Matching}

\author{%
  Alp Eren~Sari \\
  Computer Vision Group\\
  University of Bern, Switzerland\\
  \texttt{alp.sari@unibe.ch} \\
  \And
  Paolo~Favaro \\
  Computer Vision Group\\
  University of Bern, Switzerland \\
  \texttt{paolo.favaro@unibe.ch} \\
}

\begin{document}

\maketitle

\begin{abstract}
We propose FlowCut, a simple and capable method for unsupervised video instance segmentation consisting of a three-stage framework to construct a high-quality video dataset with pseudo labels. To our knowledge, our work is the first attempt to curate a video dataset with pseudo-labels for unsupervised video instance segmentation.
In the first stage, we generate pseudo-instance masks by exploiting the affinities of features from both images and optical flows. In the second stage, we construct short video segments containing high-quality, consistent pseudo-instance masks by temporally matching them across the frames. In the third stage, we use the YouTubeVIS-2021 \cite{vis2021} video dataset to extract our training instance segmentation set, and then train a video segmentation model.  FlowCut achieves state-of-the-art performance on the YouTubeVIS-2019 \cite{Yang2019vis}, YouTubeVIS-2021 \cite{vis2021}, DAVIS-2017 \cite{pont20172017}, and DAVIS-2017 Motion \cite{xie2022segmenting} benchmarks.
\end{abstract}

\section{Introduction}
\label{sec:intro}
Video instance segmentation is an important computer vision application that could be used in various scenarios, including video surveillance, autonomous driving, and video editing. These applications rely on accurately identifying and tracking individual objects across videos. However, obtaining high-quality labels for video datasets is considerably costly, time-consuming, and labor-intensive, which is a major challenge. At this point, unsupervised video instance segmentation offers a convenient way for video instance segmentation without the burden of the label acquisition process.


\begin{figure*}[t]
  \centering
  \includegraphics[width=\linewidth]{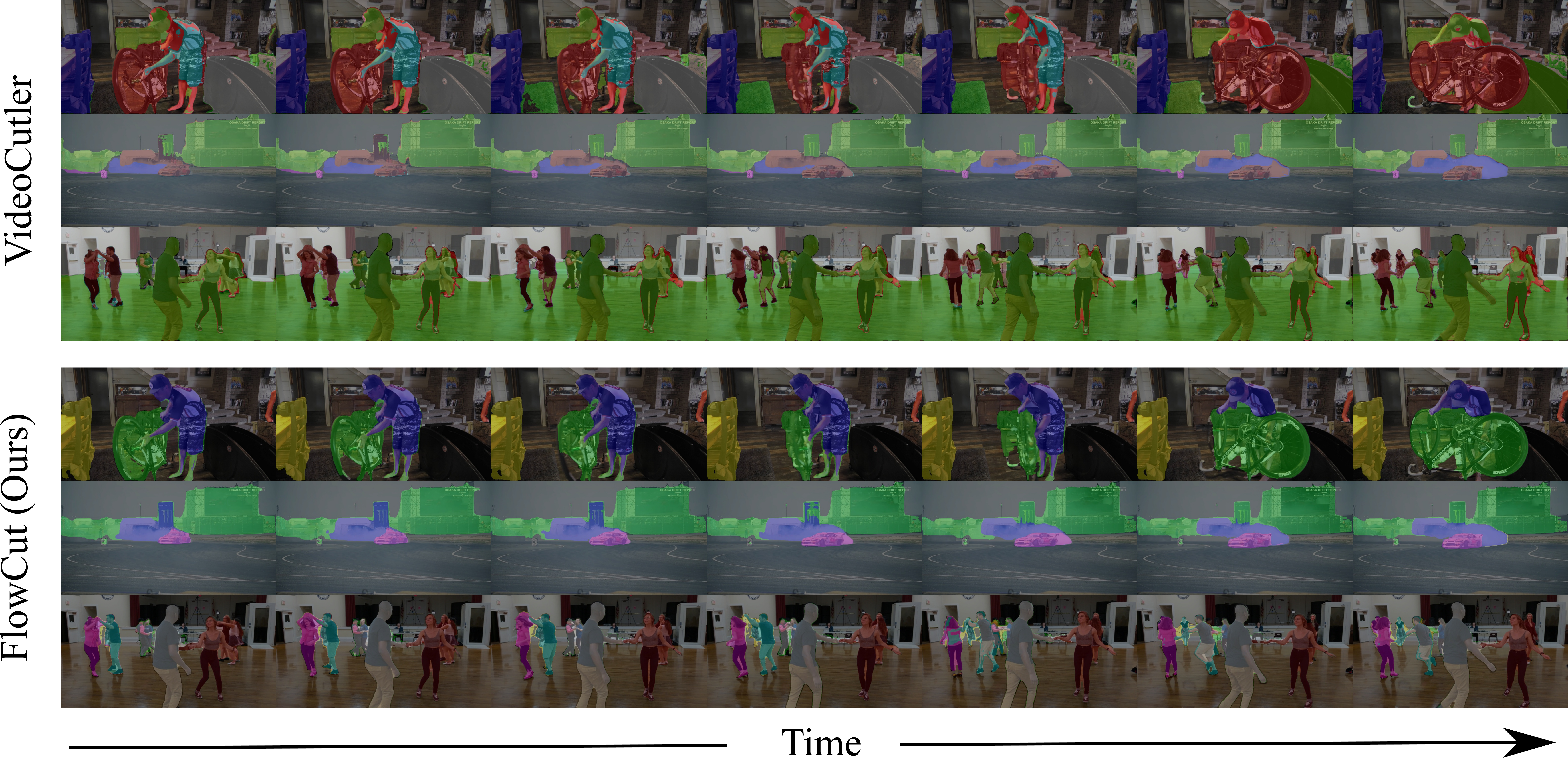}
  \caption{Qualitative comparison of VideoCutler~\cite{wang2024videocutler} and our method FlowCut on 3 videos from DAVIS-2017~\cite{pont20172017} validation set. In the first video, we can accurately detect/track both the bicycle and the cycler. In the second video, we can estimate the moving car's boundary more accurately thanks to the optical flow signal guiding our training. In the third video, we can distinguish different people dancing in the hall better.}
  \label{fig:flowcut_results}
\end{figure*}

A significant portion of the prior work in instance segmentation learning leverages optical flow of videos \cite{yang2021self, xie2022segmenting, choudhury2022guess, ye2022deformable, singh2023locate, safadoust2023multi}. While these methods have achieved notable success, they still suffer from several limitations. For instance, approaches that rely on motion models to estimate instance masks \cite{choudhury2022guess, safadoust2023multi} can struggle in scenarios where the motion model fails to generalize, negatively impacting training. Additionally, video frames in which objects exhibit minimal or no movement can further degrade model performance.

Another popular strategy is to utilize self-supervised learning (SSL) for representation learning \cite{caron2021emerging, oquab2023dinov2, aydemir2023self, wang2024videocutler}. However, SSL-based methods tend to focus on salient objects, making it challenging to accurately segment moving objects. LOCATE \cite{singh2023locate} attempts to combine the strengths of optical flow-based and SSL-based approaches, but it is limited to saliency segmentation. In contrast, our proposed method is designed to address the shortcomings of prior works by improving multi-instance segmentation while effectively leveraging motion cues and self-supervised representations.

\begin{figure}[t]
  \centering
  \includegraphics[width=\linewidth]{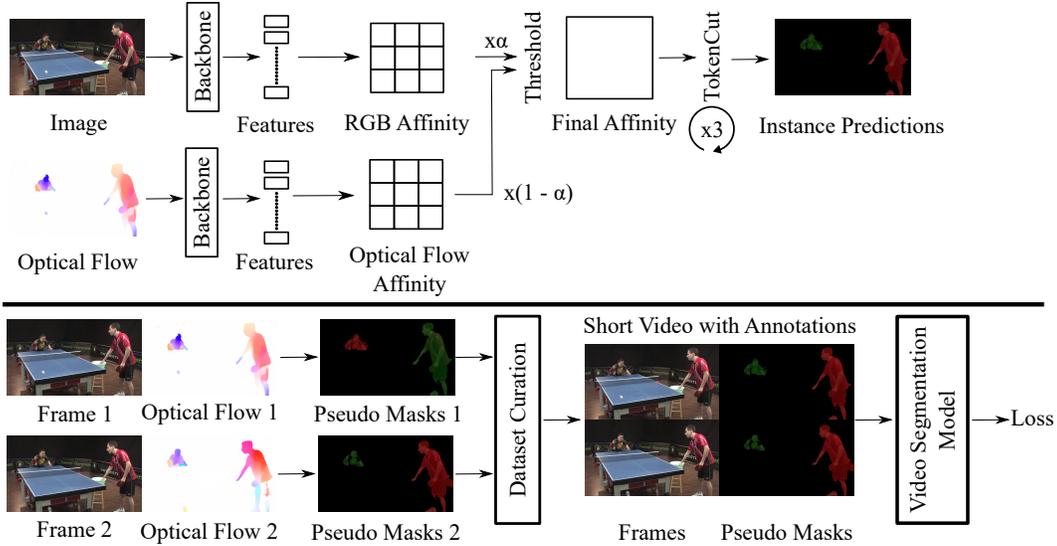}
  \caption{Illustration of the proposed synthetic image pair and corresponding pseudo-masks construction method. Top: We extract the pseudo-masks from each separate frame of a video. Bottom: We match the pseudo-masks across two frames (up to 4 time steps apart).}
  \label{fig:main_flowcut}
\end{figure}

We propose FlowCut, the first model to estimate pseudo-labels for videos for unsupervised video instance segmentation. It combines the strengths of the optical flow based and the SSL-based prior works, capable of multi-instance segmentation. FlowCut is trained on videos from YouTubeVIS-2021 \cite{vis2021} and consists of three major steps illustrated in \cref{fig:main_flowcut}. 

First, we estimate pseudo instance masks of each video frame, with the help of the corresponding optical flow. We feed the input frame and the visualization of the optical flow to a DINO backbone~\cite{caron2021emerging} to compute corresponding image features. Then, we compute two different affinity matrices with the pairwise cosine similarities of the RGB and the optical flow features and take their convex combination. Afterwards, the final affinity is computed via thresholding as proposed in TokenCut \cite{wang2023tokencut}. We apply this procedure iteratively as proposed in CutLer \cite{wang2023cut} to get multiple instance masks for each video frame.
Second, we curate a dataset of small videos (see \cref{fig:mask_matching}) using the video frames and their pseudo instance masks computed in the previous step. We compute the pairwise Intersection-over-Union (IoU) values between consecutive video frame instances, and match them based on IoU. The masks are discarded if they do not have 50\% or higher IoU with another mask. 
Third, we train the VideoMask2Former \cite{cheng2021mask2former} model on our curated dataset to get the final model.

FlowCut achieves state-of-the-art performance on  YouTubeVIS-2019 \cite{Yang2019vis}, YouTubeVIS-2021 \cite{vis2021}, DAVIS-2017 \cite{pont20172017}, and DAVIS-2017 Motion \cite{xie2022segmenting},
although it is trained with significantly less compute compared to the latest state-of-the-art method, VideoCutler~\cite{wang2024videocutler}, we achieve superior performance as illustrated in \cref{fig:flowcut_results}. 
In summary, our contributions are
\begin{itemize}
    \item To the best of our knowledge, our method is the first to estimate pseudo-labels for video datasets for unsupervised video instance segmentation;
    \item We introduce a novel pseudo mask computation and dataset curation technique, which results in high-quality segmentation targets for video instance segmentation; 
    \item Our techniques can be applied easily to new video datasets and integrated with other image datasets with limited effort. The code will be published upon publication;
    \item We achieve new state-of-the-art on YouTubeVIS-2019 \cite{Yang2019vis}, YouTubeVIS-2021 \cite{vis2021}, DAVIS-2017 \cite{pont20172017}, and DAVIS-2017 Motion \cite{xie2022segmenting} for unsupervised video instance segmentation.
\end{itemize}

\section{Related Work}
\label{sec:related_work}


\par\noindent\textbf{Supervised segmentation} methods have been developed in a variety of deep architectures \cite{ronneberger2015u, chen2017rethinking, cheng2021per, qin2020u2} and have recently demonstrated astonishing performance \cite{kirillov2023segment, ke2024segment, ravi2024sam} on both images and videos. However, because supervision requires a lot of human labor to obtain accurate segmentation masks, these methods face remarkable challenges when scaling to large datasets.
\par\noindent\textbf{Self-supervised representations (SSL)} have become a vital part of computer vision, following the success of several pioneering works \cite{chen2020simple, he2020momentum, caron2020unsupervised, caron2021emerging, he2022masked}. Among these methods, DINO \cite{caron2021emerging} and DINOv2 \cite{oquab2023dinov2} are widely used due to their high performance in various downstream tasks.
\par\noindent\textbf{Spectral Methods.} \textit{Normalized cuts} \cite{shi2000normalized} demonstrated the effectiveness of spectral methods for image segmentation. \textit{Normalized cuts} have been widely adapted since their publication; however, constructing affinity matrices to obtain high-quality segmentations remains a challenge.
\par\noindent\textbf{Unsupervised image segmentation} literature commonly leverages DINO features in different ways \cite{simeoni2021localizing, melas2022deep, wang2023tokencut, simeoni2023unsupervised, shin2022unsupervised, ravindran2023sempart, sari2024boosting}. \cite{bielski2019emergence, bielski2022move} follows a different approach and exploits the ``shiftability'' of the target objects to segment, but suffers from complex adversarial training. However, all of these methods only offer foreground/background segmentation. STEGO \cite{hamilton2022unsupervised, zadaianchuk2022unsupervised, seitzer2022bridging} explored unsupervised semantic segmentation using DINO features. Cutler \cite{wang2023cut} and Cuvler \cite{arica2024cuvler} extend DINO-based segmentation into instance segmentation. UnSAM \cite{wang2025segment} extended Cutler \cite{wang2023cut} to perform promptable instance segmentation.
\par\noindent\textbf{Unsupervised video segmentation} is a challenging task since it involves detecting and tracking objects without supervision. Several works \cite{keuper2015motion, keuper2018motion, yang2021self, xie2022segmenting, choudhury2022guess, ye2022deformable, singh2023locate, safadoust2023multi} explore leveraging optical flow since it is a strong signal that can guide segmentation, especially in cases where the motion pattern of the objects significantly differs from the background.
\cite{xie2022segmenting, safadoust2023multi} extends the usage of optical flow into video instance segmentation. \cite{aydemir2023self} takes a different approach and explores self-supervised object-centric learning. VideoCutler \cite{wang2024videocutler} extends Cutler \cite{wang2023cut} to video instance segmentation by constructing artificial videos from ImageNet \cite{deng2009imagenet} with pseudo-masks and training a video segmentation model. However, the curated dataset consists of out-of-distribution, unrealistic videos, and is constructed from single images. LOCATE \cite{singh2023locate} attempts to utilize both the image and the optical flow signals for video saliency segmentation for foreground and background segmentation. In our work, we aim to mitigate the above issues.

\section{FlowCut}
\label{sec:method}
Our method focuses on the challenging task of building a high-quality segmentation dataset in an unsupervised fashion. The performance of a segmentation model trained with it will depend very much on the consistency and detail of the extracted segmentation masks.
We split the whole procedure into a three-stage pipeline.
\begin{enumerate}
    \item Initial candidate pseudo masks are extracted from each frame of a video dataset, such as YTVIS 2021 \cite{vis2021};
    \item The training samples are synthetically created by pairing frames from a video that are up to 4 frames apart; the paired frames define the input, while the targets consists of the segmentations obtained by matching the pseudo masks via their intersection-over-union (IoU) across the two frames; 
    \item Finally, a video segmentation model is trained on the artificial data obtained in the previous step.
\end{enumerate}

\subsection{Preliminaries}
First, let us introduce the unsupervised learning techniques that will be used to extract the pseudo-masks. 
\par\noindent\textbf{TokenCut}~\cite{wang2023tokencut} proposes to utilize normalized cuts with an affinity matrix constructed using cosine similarities between DINO features. The $ij$-th element of their affinity matrix is
\begin{align}
w_{ij} = \left\{
\begin{array}{l}
    1, \text{when } \langle h_{i}, h_{j}\rangle > \tau \\
    \epsilon,  \text{otherwise},
\end{array}\label{ncutaff}
\right.
\end{align}
where $h_{i}$ is the normalized DINO feature \cite{caron2021emerging,oquab2023dinov2} corresponding to the image patch $v_i$, with $i=1,\dots,N$, where $N$ is the number of image patches, $\tau$ is the threshold value and $\epsilon>0$ is a value to keep the graph fully connected.

After defining the affinity matrix, generalized eigenvalue decomposition is done for the expression 
\begin{equation}
    (D - W)x = \lambda D x
\end{equation}
and $D = \sum_j w_{ij}$, which are proposed by Normalized Cuts \cite{shi2000normalized}. The eigenvector corresponding to the second lowest eigenvalue is taken is binarized by comparing all the values to the mean of them. 
\par\noindent\textbf{LOCATE}~\cite{singh2023locate} extends the idea of TokenCut~\cite{wang2023tokencut} into the video domain by injecting the optical flow signal into the affinity matrix. Their proposed affinity matrix is given as 
\begin{align}
\label{eq:affinity_matrix}
w_{ij} = \begin{cases}
\begin{array}{l}
    1, \text{when }\alpha \langle h^\text{rgb}_{i}, h^\text{rgb}_{j}\rangle + (1 - \alpha) \langle h^\text{of}_{i}, h^\text{of}_{j}\rangle > \tau \\
    \epsilon, \text{otherwise},
\end{array}
\end{cases}
\end{align}
where $w_{ij}$ is the $ij$-th element of the affinity matrix, $h^\text{rgb}_{i}$ is the normalized DINO feature corresponding to the image patch $i$, $h^\text{of}_{i}$ is the normalized DINO feature corresponding to the optical flow visualization patch $i$, and $\tau$ and $\epsilon$ are the same as in TokenCut~\cite{wang2023tokencut}. 
\par\noindent\textbf{Cutler}~\cite{wang2023cut} proposes to utilize TokenCut iteratively to perform instance segmentation. 
Their proposed affinity matrix is the same as in TokenCut, but is updated iteratively after each instance mask prediction to remove the affinity entries corresponding to the previous selected masks via 
\begin{align}
\label{eq:cutler}
w_{ij}^{t+1} = w_{ij} \cdot \left(\prod_{s=1}^{t} \hat{M}_{i}^{s} \right) \cdot \left(\prod_{s=1}^{t} \hat{M}_{j}^{s} \right),
\end{align}
where $w_{ij}$ is the initial affinity constructed using the DINO features, $\hat{M}^{s}$ is the inverted and vectorized (flattened) version of the mask $M^{s}$ predicted at the $s$-th iteration, i.e., $\hat{M}^{s} \doteq \text{vect}(1 - M^{s})$.
\subsection{Pseudo-Mask Estimation}
\label{sec:method_flowcut}
We propose to combine the affinity formulation of LOCATE~\cite{singh2023locate} and the iterative segmentation scheme of Cutler~\cite{wang2023cut} to estimate instance masks of videos as illustrated in \cref{fig:main_flowcut}. We begin with the initial affinity matrix in \cref{eq:affinity_matrix}, 
where 
\begin{align}
    h^\text{rgb}_{1:N} &= f(I), \\
    h^\text{of}_{1:N} &= f(I_{of}).
\end{align}
$f$ is the DINO backbone used to extract image features, $I$ is the input frame, $I_\text{of}$ is the \emph{visualization} of the optical flow between the input frame and a consecutive frame. We update this affinity to extract the pseudo-masks in an iterative way as in \cref{eq:cutler}.

\begin{figure*}[t]
  \centering
  \includegraphics[width=\linewidth]{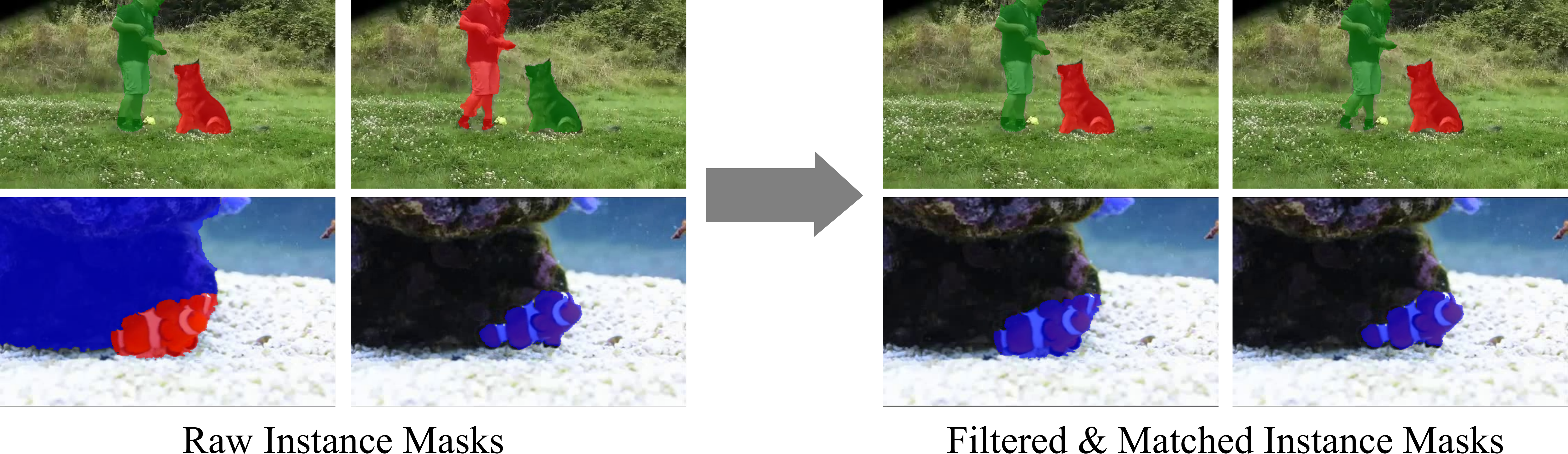}
  \caption{Illustration of the proposed mask-matching algorithm. Top row: If the mask indexing across the two frames is inconsistent, the IoUs can be used to rectify the correspondences. Bottom row: If masks are missing in one of the 2 frames, they are considered unreliable, and the IoUs can also be used to remove them. }
  \label{fig:mask_matching}
\end{figure*}

\subsection{Automated Dateset Curation}
\label{sec:method_dataset_curation}
After generating pseudo-instance masks, it is crucial to establish correspondences between instances across frames to construct short video clips with consistent pseudo-masks. This step enables the effective training of a video segmentation model. To achieve this, we adopt a straightforward approach by creating two-frame video segments from all available frames and their corresponding pseudo-masks. However, this process presents two key challenges. First, the same instance may not always be detected across consecutive frames. For instance, in a video with a moving bear, the bear may only be identified in one of the frames. Second, even when instances are detected in both frames, their indexing may be inconsistent across frames. For example, if a bear and a dog appear in two consecutive frames, their per-instance order might be ``bear-dog'' in the first frame and ``dog-bear'' in the second. \\
To address these challenges, we propose a filtering algorithm that follows these steps:
\begin{enumerate}
    \item Compute the Intersection-over-Union (IoU) matrix between all instances in the first and second frames;
    \item For each instance in the first frame, select the instance with the highest IoU in the second frame, and match the instances if the IoU is higher than 50\%; 
    \item Repeat the previous step for all instances in the first frame to ensure optimal instance matching across frames.
\end{enumerate}
The proposed mask-matching algorithm is illustrated in \cref{fig:mask_matching}, and the whole instance filtering and matching process is described in detail in \cref{algo:instance_matching} in Pytorch style.  
\begin{algorithm}
\caption{Dataset Curation}
\begin{algorithmic}[1]
 \Require The instances $I_1$ and $I_2$ of frames $F_1$ and $F_2$
\State iou\_matrix = zeros($|I_1|$, $|I_2|$) \# IoU matrix 
\For{$i$ in range($|I_1|$)}
    \For{$j$ in range($|I_2|$)}
    \State \#  Compute IoU between instance pair
        \State $iou\_matrix[i, j] = IoU(I_1[i], I_2[j])$ 
    \EndFor
\EndFor
\State $I_1^{new}$, $I_2^{new}$ = [], []
\For{$i$ in range($|I_1|$)}
    \State $j$ = argmax(M[i, :]) 
    \If{$iou\_matrix[i, j] > 0.5$}
        \State $I_1^{new}$.append($I_1[i]$)
        \State $I_2^{new}$.append($I_2[j]$)
    \EndIf
\EndFor
\State flag = len($I_1^{new}$) $>$ 0
\State \Return  $I_1^{new}$, $I_2^{new}$, flag
\end{algorithmic}
\label{algo:instance_matching}
\end{algorithm}

\subsection{ Video Segmentation Model}
\label{sec:method_video_segmentation}
After the steps described in \cref{sec:method_flowcut} and \cref{sec:method_flowcut}, we have two frame length videos with consistent instance masks, which may be used to train a video segmentation model. \\ 
We follow a similar approach to \cite{wang2024videocutler} to distill the dataset we constructed and utilize VideoMask2Former~\cite{cheng2021mask2former} with a ResNet~50 \cite{he2016deep} backbone, which is a powerful model that can segment and track multiple instances simultaneously.

\section{Experimental Results}
\label{sec:experimental_results}



\begin{table*}[t]
    \centering
    \caption{Unsupervised multi-instance video segmentation results on DAVIS-2017 \cite{pont20172017}. VideoCutler$^\ast$ denotes that evaluation is done with the pretrained checkpoint provided by the authors for a fair comparison. The best results are indicated \textbf{bold.}} 
    \label{tab:results_davis2017} \vspace{4pt}
    \begin{tabular*}{\textwidth}{@{\extracolsep{\fill}}l|l|ccccccc@{}}
        \hline
        \multirow{2}{*}{Methods} & \multirow{2}{*}{Flow} & \multicolumn{3}{c}{DAVIS-2017 \cite{pont20172017}} \rule{0pt}{10pt} & \multicolumn{3}{c}{DAVIS-2017-Motion \cite{xie2022segmenting}} \\ 
        \cline{3-8} & & $\mathcal{J}\&\mathcal{F}$ \rule{0pt}{10pt} & $\mathcal{J}$ & $\mathcal{F}$ & $\mathcal{J}\&\mathcal{F}$ & $\mathcal{J}$ & $\mathcal{F}$ \\ \hline
        MotionGroup \cite{yang2021self} \rule{0pt}{10pt} & \checkmark & - & - & - & 39.5 & 44.9 & 34.2 \\
        OCLR \cite{xie2022segmenting} & \checkmark & 39.6 & 38.2 & 41.1 & 55.1 & 54.5 & 55.7 \\
        Safadoust and Güney \cite{safadoust2023multi} & \checkmark & - & - & - & 55.3 & \textbf{55.3} & 55.3 \\
        VideoCutLER$^\ast$~\cite{wang2024videocutler} & \texttimes & 42.4 & 39.2 & \textbf{45.6} & 54.6 & 53.3 & 55.9 \\
        FlowCut & \checkmark & \textbf{43.5} & \textbf{41.7} & 45.2 & \textbf{58.3} & 53.3 & \textbf{57.5} \\ 
        \hline
    \end{tabular*}
\end{table*}
We have conducted extensive experiments on several benchmarks to verify the effectiveness of our method. We report unsupervised video instance segmentation results in \cref{sec:experiments_instance_segmentation} for segmentation/tracking and boundary accuracy. We also performed an ablation study reported in \cref{sec:experiments_ablations} which proves the effectiveness of fusing optical flow information. Eventually, we also discuss the limitations of our method in \cref{sec:experiments_limitation}.

\subsection{Implementation Details}
\label{sec:experiments_implementation}
\par\noindent\textbf{FlowCut.} We employ RAFT \cite{teed2020raft} to estimate optical flow between consecutive frames with a timestamp difference of $+4$ or $-4$. The DINO ViT-Small/16 model \cite{caron2021emerging} is employed to extract image and optical flow features. Each image is resized to $360 \times 640$ pixels, and the maximum instance masks are set to 3, and the threshold ($\tau$) for constructing the affinity matrix is set to 0.15. Additionally, Conditional Random Fields (CRFs) \cite{krahenbuhl2011efficient} are applied as a post-processing step to refine mask boundaries and improve segmentation accuracy.
\par\noindent\textbf{Training Settings.} We employ the VideoMask2Former \cite{cheng2021mask2former} model with ResNet 50 \cite{he2016deep} backbone initialized with Cutler \cite{wang2023cut} weights. We set the initial learning rate to $1 \times {10}^{-6}$ with AdamW \cite{loshchilov2017decoupled}, batch size to 4, dropout rate of decoders to 0.3, and train our model for 5000 iterations on a single 3090 TI GPU with 24 GB memory. We resize the image into $360 \times 640$ and apply random flipping, brightness, contrast, and rotation changes to the input frames as part of our data augmentation process. The same transformations, except for the color transformations, are also applied to the instance masks to maintain consistency. We train our model for 80000 iterations for the DAVIS-2017 \cite{pont20172017} and DAVIS-2017 Motion \cite{xie2022segmenting} experiments in \cref{tab:results_davis2017}.
\par\noindent\textbf{Training Data.} Our method is trained on a dataset of 167,365 two-frame short video segments, which are generated from the frames of the YouTubeVIS-2021 training set \cite{vis2021}, consisting of 2,985 videos. These short videos are constructed using frames with timestamp differences of 1, 2, 3, and 4, while the corresponding pseudo-masks are generated following the procedures introduced in \cref{sec:method_flowcut} and \cref{sec:method_dataset_curation}. For DAVIS-2017 \cite{pont20172017} and DAVIS-2017 Motion \cite{xie2022segmenting} experiments, we add randomly selected 50000 images from ImageNet~\cite{deng2009imagenet} and their pseudo labels computed by the procedure introduced in \cite{wang2024videocutler}. Our backbone DINO~\cite{caron2021emerging} is trained on ImageNet~\cite{deng2009imagenet}, and the boundary accuracy is superior in this dataset. We find that this significantly improves boundary quality in DAVIS-2017 and DAVIS-2017 Motion benchmarks.
\par\noindent\textbf{Test Data.} We evaluate our model on the training sets of YouTubeVIS-2019 \cite{Yang2019vis} (2,238 videos) and YouTubeVIS-2021 \cite{vis2021} (2,985 videos), as well as on the validation sets of DAVIS-2017 \cite{pont20172017} and the motion labels introduced by \cite{xie2022segmenting}. It is important to note that while the YouTubeVIS-2021 training set is used for both training and evaluation, the evaluation remains unsupervised, as our method relies solely on pseudo-masks during training.
\par\noindent\textbf{Evaluation Settings.} We used a score threshold of 0.8 for instance detection on YouTubeVIS-2019 \cite{Yang2019vis} and YouTubeVIS-2021 \cite{vis2021} datasets. We set this value to 0.3 for DAVIS-2017 \cite{pont20172017} and DAVIS-2017-Motion \cite{xie2022segmenting} evaluations. 
\par\noindent\textbf{Evaluation Metrics.} The YouTubeVIS-2019 \cite{Yang2019vis} and YouTubeVIS-2021 \cite{vis2021} datasets are evaluated with Average Precision (AP) and Average Recall (AR) metrics in a class agnostic way, in which detections or misses are decided based on the IoU between the predicted and the ground-truth instances. 
\par For DAVIS-2017 \cite{pont20172017} and DAVIS-2017 Motion \cite{xie2022segmenting}, we employ their official metrics which are the region measure ($\mathcal{J}$), the boundary measure ($\mathcal{F}$) and their average ($\mathcal{J}\&\mathcal{F}$). The predicted instances and the ground-truth instances are matched with Hungarian matching.


\begin{table*}[t]
    \centering
    \caption{Unsupervised multi-instance video segmentation results on the training set of YouTubeVIS-2019 \cite{Yang2019vis}. We achieve state-of-the-art results on all metrics except $AP_L$. VideoCutler$^\ast$ denotes that evaluation is done with the pretrained checkpoint provided by the authors for a fair comparison. The best results are indicated \textbf{bold.}}
    \label{tab:ar_ap_ytvis19} \vspace{4pt}
    \begin{tabular*}{\textwidth}{@{\extracolsep{\fill}}l|l|cccccccc}
        \hline
        Methods \rule{0pt}{10pt} & Flow & AP$_{50}$ & AP$_{75}$ & AP & AP$_S$ & AP$_M$ & AP$_L$ & AR$_{10}$ \\ \hline
        MotionGroup~\cite{yang2021self} \rule{0pt}{10pt} & \checkmark & 1.3 & 0.1 & 0.3 & 0.2 & 0.3 & 0.5 & 1.7 \\
        OCLR~\cite{xie2022segmenting} & \checkmark & 4.8 & 0.4 & 1.3 & 0.0 & 1.2 & 5.5 & 11.0 \\
        CutLER~\cite{wang2023cut} & \texttimes & 37.5 & 14.6 & 17.1 & 3.3 & 13.9 & 27.6 & 30.4 \\
        VideoCutLER$^\ast$~\cite{wang2024videocutler} & \texttimes & 47.0 & 22.1 & 23.8 & 5.5 & 20.4 & \textbf{33.8} & 38.8 \\
        FlowCut (Ours) & \checkmark & \textbf{47.9} & \textbf{22.8} & \textbf{24.3} & \textbf{7.1} & \textbf{21.8} & \textbf{33.8} & \textbf{39.1} \\ \hline
        \textit{vs. prev. SOTA} \rule{0pt}{10pt} & & \textcolor{green}{+0.9} & \textcolor{green}{+0.7} & \textcolor{green}{+0.5} & \textcolor{green}{+1.6} & \textcolor{green}{+1.4} & \textcolor{green}{0.0} & \textcolor{green}{+0.3} \\
        \hline
    \end{tabular*}
\end{table*}

\begin{table*}[t]
    \centering
    \caption{Unsupervised multi-instance video segmentation results on the training set of YouTubeVIS-2021 \cite{vis2021}. We achieve state-of-the-art results on all metrics with significant margins. VideoCutler$^\ast$ denotes that evaluation is done with the pretrained checkpoint provided by the authors for a fair comparison. The best results are indicated \textbf{bold.}}
    \label{tab:ar_ap_ytvis21} \vspace{4pt}
    \begin{tabular*}{\textwidth}{@{\extracolsep{\fill}}l|l|cccccccc}
        \hline
        Methods \rule{0pt}{10pt} & Flow & AP$_{50}$ & AP$_{75}$ & AP & AP$_S$ & AP$_M$ & AP$_L$ & AR$_{10}$ \\ \hline
        MotionGroup~\cite{yang2021self} \rule{0pt}{10pt} & \checkmark & 1.1 & 0.1 & 0.2 & 0.1 & 0.2 & 0.5 & 1.5 \\
        OCLR~\cite{xie2022segmenting} & \checkmark & 4.4 & 0.3 & 1.2 & 0.1 & 1.6 & 7.1 & 9.6 \\
        CutLER~\cite{wang2023cut} & \texttimes & 29.2 & 10.4 & 12.8 & 3.1 & 12.8 & 27.8 & 22.6 \\
        VideoCutLER$^\ast$~\cite{wang2024videocutler} & \texttimes & 36.1 & 15.4 & 17.4 & 5.2 & 17.7 & 33.0 & 28.9 \\
        FlowCut (Ours) & \checkmark & \textbf{37.4} & \textbf{16.1} & \textbf{18.0} & \textbf{5.8} & \textbf{19.2} & \textbf{33.4} & \textbf{29.7} \\ \hline
        \textit{vs. prev. SOTA} \rule{0pt}{10pt} & & \textcolor{green}{+1.3} & \textcolor{green}{+0.7} & \textcolor{green}{+0.6} & \textcolor{green}{+0.6} & \textcolor{green}{+1.5} & \textcolor{green}{+0.4} & \textcolor{green}{+0.8} \\
        \hline
    \end{tabular*}
\end{table*}

\subsection{Unsupervised Video Instance Segmentation}
\label{sec:experiments_instance_segmentation}
We present the unsupervised segmentation performance of our method on the training sets of YouTubeVIS-2019~\cite{Yang2019vis} and YouTubeVIS-2021~\cite{vis2021} in \cref{tab:ar_ap_ytvis19} and \cref{tab:ar_ap_ytvis21}, respectively. We also report the DAVIS-2017 \cite{pont20172017} and the DAVIS-2017 Motion \cite{xie2022segmenting} evaluations in \cref{tab:results_davis2017}. FlowCut achieves SotA for $\mathcal{J}\&\mathcal{F}$ on all benchmarks, for $\mathcal{J}$ on DAVIS-2017, and for $\mathcal{F}$ on DAVIS-2017 Motion. Our approach demonstrates consistent improvements on all datasets, achieving a new state-of-the-art (SotA) performance in unsupervised video instance segmentation. We also provide a qualitative comparison of VideoCutler~\cite{wang2024videocutler} and our method, FlowCut, on sampled videos from the DAVIS-2017~\cite{pont20172017} validation set in \cref{fig:flowcut_results}. This comparison also illustrates the strength of our method in detecting and tracking independent moving objects in complex scenes. We evaluate VideoCutler \cite{wang2024videocutler} using the pretrained checkpoint provided by the authors for a fair comparison.

\begin{table*}[t]
    \centering
    \caption{Ablations results on the effect of optical flow information on YouTubeVIS-2019~\cite{Yang2019vis} train set. We show that performance is improved consistently with optical flow usage during the pseudo-mask estimation. Specifically, we get the largest improvement when we train the model on the dataset curated from YouTubeVIS-2019~\cite{Yang2019vis}. The best results are indicated \textbf{bold.}}
    \label{tab:ablation_ytvis19} \vspace{4pt}
    \begin{tabular*}{\textwidth}{@{\extracolsep{\fill}}l|l|cccccccc}
        \hline
        \multirow{2}{*}{Training Dataset} & \multirow{2}{*}{Flow} & \multicolumn{7}{c}{YouTubeVIS-2019 \cite{Yang2019vis}} \rule{0pt}{10pt} \\ 
        \cline{3-9} & & AP$_{50}$ \rule{0pt}{10pt} & AP$_{75}$ & AP & AP$_S$ & AP$_M$ & AP$_L$ & AR$_{10}$ \\ \hline
       \multirow{2}{*}{YouTubeVIS-2019 \cite{Yang2019vis}} & \texttimes \rule{0pt}{10pt} & 43.0 & 20.7 & 22.0 & 3.4 & 17.0 & 33.3 & 34.8 \\
        & \checkmark & \textbf{49.6} & \textbf{23.4} & \textbf{25.1} & 4.3 & 18.1 & \textbf{38.3} & 38.5 \\
        \cline{2-9} & \textit{Improvement} \rule{0pt}{10pt} & \textcolor{green}{+6.6} & \textcolor{green}{+2.7} & \textcolor{green}{+3.1} & \textcolor{green}{+0.9} & \textcolor{green}{+1.1} & \textcolor{green}{5.0} & \textcolor{green}{+3.7} \\
        \hline
        \multirow{2}{*}{YouTubeVIS-2021 \cite{vis2021}} & \texttimes \rule{0pt}{10pt} & 43.4 & 21.3 & 22.4 & 5.7 & 19.5 & 31.8 & 37.6 \\
        & \checkmark & 47.9 & 22.8 & 24.3 & \textbf{7.1} & \textbf{21.8} & 33.8 & \textbf{39.1} \\
        \cline{2-9} & \textit{Improvement} \rule{0pt}{10pt} & \textcolor{green}{+4.5} & \textcolor{green}{+1.5} & \textcolor{green}{+1.9} & \textcolor{green}{+1.4} & \textcolor{green}{+2.3} & \textcolor{green}{2.0} & \textcolor{green}{+1.5} \\
        \hline
    \end{tabular*}
\end{table*}

\begin{table*}[t]
    \centering
    \caption{Ablations results on the effect of optical flow information on YouTubeVIS-2021 \cite{vis2021} train set.  We show that performance is improved consistently with optical flow usage during the pseudo-mask estimation. The model trained with the dataset curated from YouTubeVIS-2021 \cite{vis2021} provides a higher performance boost. The best results are indicated \textbf{bold.}}
    \label{tab:ablation_ytvis21} \vspace{4pt}
    \begin{tabular*}{\textwidth}{@{\extracolsep{\fill}}l|l|cccccccc}
        \hline
        \multirow{2}{*}{Training Dataset} & \multirow{2}{*}{Flow} & \multicolumn{7}{c}{YouTubeVIS-2021\cite{vis2021}} \rule{0pt}{10pt} \\ 
        \cline{3-9} &  & AP$_{50}$ \rule{0pt}{10pt} & AP$_{75}$ & AP & AP$_S$ & AP$_M$ & AP$_L$ & AR$_{10}$ \\ \hline
        \multirow{2}{*}{YouTubeVIS-2019 \cite{Yang2019vis}} & \texttimes \rule{0pt}{10pt} & 32.6 & 14.4 & 16.0 & 3.2 & 15.1 & 32.9 & 25.4 \\
         & \checkmark & 35.2 & 14.6 & 16.7 & 3.7 & 16.0 & \textbf{33.9} & 26.2 \\
        \cline{2-9} & \textit{Improvement} \rule{0pt}{10pt} & \textcolor{green}{+2.6} & \textcolor{green}{+0.2} & \textcolor{green}{+0.7} & \textcolor{green}{+0.5} & \textcolor{green}{+0.9} & \textcolor{green}{+1.0} & \textcolor{green}{+0.8} \\
        \hline
        \multirow{2}{*}{YouTubeVIS-2021 \cite{vis2021}} & \texttimes \rule{0pt}{10pt} & 32.9 & 14.8 & 16.2 & 4.6 & 16.6 & 31.1 & 28.0 \\
         & \checkmark & \textbf{37.4} & \textbf{16.1} & \textbf{18.0} & \textbf{5.8} & \textbf{19.2} & 33.4 & \textbf{29.7} \\
        \cline{2-9} & \textit{Improvement} \rule{0pt}{10pt} & \textcolor{green}{+4.5} & \textcolor{green}{+1.3} & \textcolor{green}{+1.8} & \textcolor{green}{+1.2} & \textcolor{green}{+2.6} & \textcolor{green}{+2.3} & \textcolor{green}{+1.7} \\
        \hline 
    \end{tabular*}
\end{table*}

\subsection{Ablations}
\label{sec:experiments_ablations}
In Table \cref{tab:ablation_ytvis19} and Table \cref{tab:ablation_ytvis21}, we present the ablations on the effect of the optical flow utilization to estimate the pseudo-masks. We simply set the $\alpha$ parameter in \cref{eq:affinity_matrix} to 1 to remove the effect of optical flow. Hence, the ablations denoted as "without optical flow" compute the affinity matrix with $\alpha=1$. We evaluate 4 different models trained with 4 different pseudo-masks computed: YouTubeVIS-2019 \cite{Yang2019vis} training set without optical flow, YouTubeVIS-2019 \cite{Yang2019vis} training set with optical flow, YouTubeVIS-2021 \cite{vis2021} training set without optical flow, YouTubeVIS-2021 \cite{vis2021} training set with optical flow.

We evaluate four different models on the YouTubeVIS-2019 \cite{Yang2019vis} and YouTubeVIS-2021 \cite{vis2021} training sets, with the results presented in \cref{tab:ablation_ytvis19} and \cref{tab:ablation_ytvis21}, respectively. A consistent trend observed in both tables is that incorporating optical flow information leads to a significant improvement in video instance segmentation performance. This highlights the importance of motion cues in accurately tracking and segmenting objects across frames. To ensure a fair comparison, all models are trained under identical conditions, following the implementation details outlined in \cref{sec:experiments_implementation}. Another critical result that can be inferred from \cref{tab:ablation_ytvis19} and \cref{tab:ablation_ytvis21} is the importance of in-domain training data. We observe that when the training data and the test data is the same, video instance segmentation improves significantly.

\subsection{Limitations and Potential Impacts}
\label{sec:experiments_limitation}
\par
\begin{figure}[bt]
  \centering
  \includegraphics[width=\linewidth]{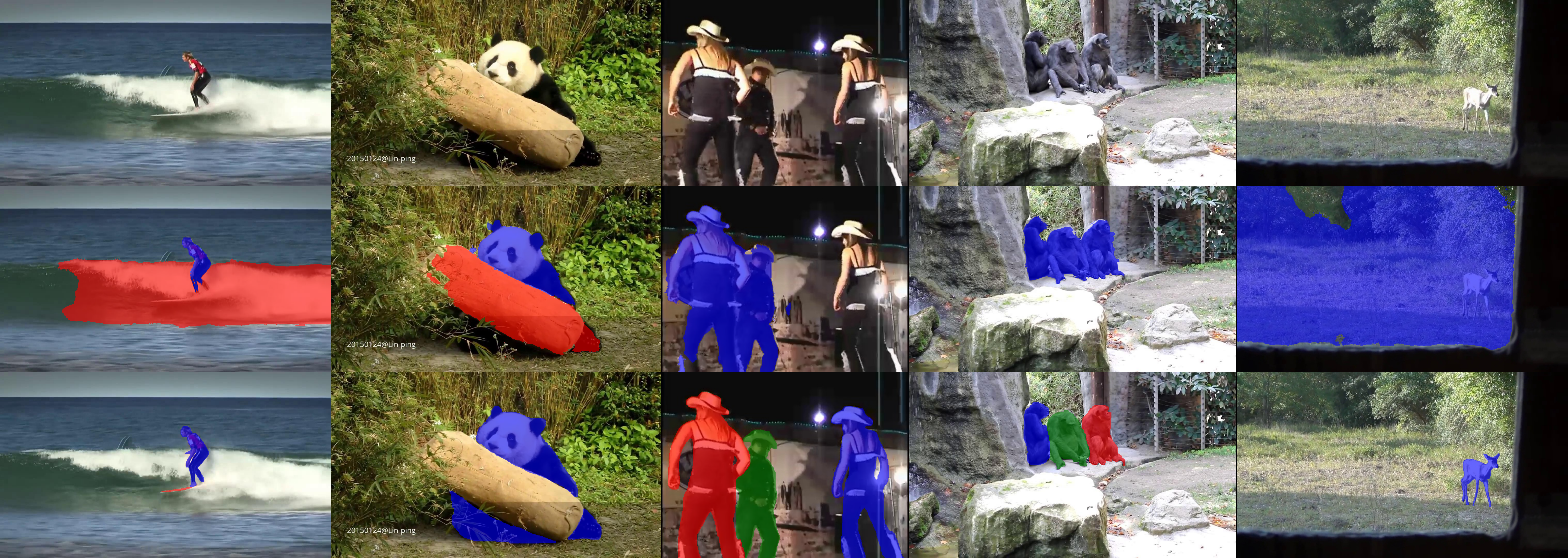}
  \caption{Some example cases of the failures from YouTubeVIS-2021 \cite{vis2021}. The first row shows the input images, the second row shows the pseudo-masks, and the third row shows the ground-truth masks.}
  \label{fig:failures}
\end{figure}

\noindent\textbf{Limitations.} The pseudo-masks estimated in \cref{sec:method_flowcut} may contain errors due to the limitations of DINO features, which are not always perfectly reliable. This presents one of the most challenging aspects of our framework. Examples of failed pseudo-masks can be seen in \cref{fig:failures}. 

One major challenge is the detection of small objects, as the spatial resolution of the extracted image features is lower than that of the original input image, making it harder to accurately segment smaller instances, as shown in the fifth image of \cref{fig:failures}. Additionally, the presence of static salient regions in the image can lead to incorrect segmentations. If an object appears visually distinct but does not move, it should not be detected or tracked; however, it may still be mistakenly included in the pseudo-mask extraction process. For instance, as shown in \cref{fig:failures}, the waves in the first image and the wood in the second image are incorrectly segmented as objects. Another significant source of error is severe occlusion, where objects are partially or fully obscured by others. The third and fourth images in \cref{fig:failures} demonstrate how occlusion can negatively impact instance segmentation, further complicating the task.
\par\noindent\textbf{Potential Positive Impacts.} Our method offers a cost-effective solution for students or research groups with limited resources to conduct research on video instance segmentation due to its unsupervised nature. We believe these aspects of our method could impact society positively.
\par\noindent\textbf{Potential Negative Impacts.} Our method is unlikely to have a negative impact on society, since we perform video instance segmentation, which can also be done by other supervised methods or even manually by humans.

\section{Conclusions and Future Work}
\label{sec:conclusions}
\par\noindent\textbf{Conclusions.} We introduce FlowCut, a simple yet powerful unsupervised video segmentation model that can be trained efficiently using modest computational resources. Our approach is the first method that estimates pseudo-labels for unsupervised video instance segmentation. It effectively integrates the strengths of various unsupervised image and video segmentation techniques, creating a more robust and adaptable framework. \\
FlowCut operates in three stages on video datasets. First, we estimate pseudo instance masks for each frame of the videos with the help of the estimated optical flow with RAFT \cite{teed2020raft} (\cref{sec:method_flowcut}). Second, we apply our proposed dataset curation procedure in \cref{sec:method_dataset_curation} to construct a target video dataset with 2-frame videos. Third, train the VideoMask2Former \cite{cheng2021mask2former} model with the curated dataset. These stages can be applied easily and in a short time to new video datasets.\\ 
Our method achieves state-of-the-art performance on the YouTubeVIS-2019 \cite{Yang2019vis}, YouTubeVIS-2021 \cite{vis2021}, DAVIS-2017 \cite{pont20172017} and the DAVIS-2017 Motion \cite{xie2022segmenting} benchmarks, demonstrating its effectiveness. Given that annotating segmentation labels for video datasets is labor-intensive and costly, unsupervised video instance segmentation provides a feasible and scalable alternative, reducing the need for extensive manual labeling while maintaining high performance.
\par\noindent\textbf{Future Work.} A promising direction for future research is leveraging entire video sequences rather than relying solely on individual frames and their optical flow to estimate pseudo-masks. Incorporating temporal information across multiple frames could enhance segmentation consistency and accuracy. 


\bibliographystyle{plain}
\bibliography{main}

\end{document}